\def\year{2022}\relax
\documentclass[letterpaper]{article} 
\usepackage{aaai22}  
\usepackage{times}  
\usepackage{helvet}  
\usepackage{courier}  
\usepackage[hyphens]{url}  
\usepackage{graphicx} 
\usepackage{natbib}  
\usepackage{caption} 
\DeclareCaptionStyle{ruled}{labelfont=normalfont,labelsep=colon,strut=off} 
\usepackage{algorithm}
\usepackage{algorithmic}

\usepackage{newfloat}
\usepackage{listings}
\floatstyle{ruled}
\newfloat{listing}{tb}{lst}{}
\floatname{listing}{Listing}
\usepackage{amsmath}
\usepackage{amssymb}
\usepackage{amsfonts}
\usepackage{multirow}
\usepackage{booktabs}

\usepackage[hidelinks]{hyperref}

\title{ALGAN: Anomaly Detection by Generating Pseudo Anomalous Data\\via Latent Variables\footnote{This paper has been accepted for publication in IEEE Access. \href{https://doi.org/10.1109/ACCESS.2022.3169594}{DOI: 10.1109/ACCESS.2022.3169594}}}
\author {
    Hironori Murase,\textsuperscript{\rm 1}
    Kenji Fukumizu, \textsuperscript{\rm 2,1}
}
\affiliations {
    \textsuperscript{\rm 1} The Graduate University for Advanced Studies, SOKENDAI\\
    \textsuperscript{\rm 2} The Institute of Statistical Mathematics\\
    h-murase@ism.ac.jp, fukumizu@ism.ac.jp
}

\begin{document}

\maketitle

\begin{abstract}
In many anomaly detection tasks, where anomalous data rarely appear and are difficult to collect, training using only normal data is important. Although it is possible to manually create anomalous data using prior knowledge, they may be subject to user bias. In this paper, we propose an \underline{A}nomalous \underline{L}atent variable \underline{G}enerative \underline{A}dversarial \underline{N}etwork (ALGAN) in which the GAN generator produces pseudo-anomalous data as well as fake-normal data, whereas the discriminator is trained to distinguish between normal and pseudo-anomalous data. This differs from the standard GAN discriminator, which specializes in classifying two similar classes. 
The training dataset contains only normal data; the latent variables are introduced in anomalous states and are input into the generator to produce diverse pseudo-anomalous data. 
We compared the performance of ALGAN with other existing methods on the MVTec-AD, Magnetic Tile Defects, and COIL-100 datasets. The experimental results showed that ALGAN exhibited an AUROC comparable to  those of state-of-the-art methods while achieving a much faster prediction time.
\end{abstract}

\section{Introduction}

Anomaly detection refers to the technique of distinguishing between unexpected and normal data and is closely related to outlier detection and novelty detection~\cite{Chandola_etal_2009}.
Practical examples include fraud detection to identify unauthorized access~\cite{rodda2016classimbalance}, medical diagnosis to discover lesion sites from medical images~\cite{schlegl2017unsupervised}, surveillance to find suspicious behavior in real-time videos~\cite{sabokrou2015real}, and optical inspection for detecting defects in industrial products~\cite{BergmannLFSS19}. A common feature of these applications is the discovery of undesirable data.

The difficulty with anomaly detection is that, in many cases, anomalous data are rarely observed and are of a wide variety; hence, the learning of anomaly detection models suffers from the difficulty of imbalanced or one-class classification. Various methods have been proposed to address this issue, such as creating a dataset containing new anomalous data not included in conventional datasets~\cite{8926471}, verifying classification performance for anomalous data that are rarely observed~\cite{rodda2016classimbalance}, and developing a new one-class classifier for complex data such as image and sequence data~\cite{chalapathy2018oneclassneuralnet}. The augmentation of anomalous data from different sources has also been proposed (e.g., out-of-distribution data~\cite{kawachi2018complementary,hendrycks2018deep}); however, in such cases, undesired biases may be introduced.

\begin{figure*}[!t]
    \includegraphics[width=\linewidth]{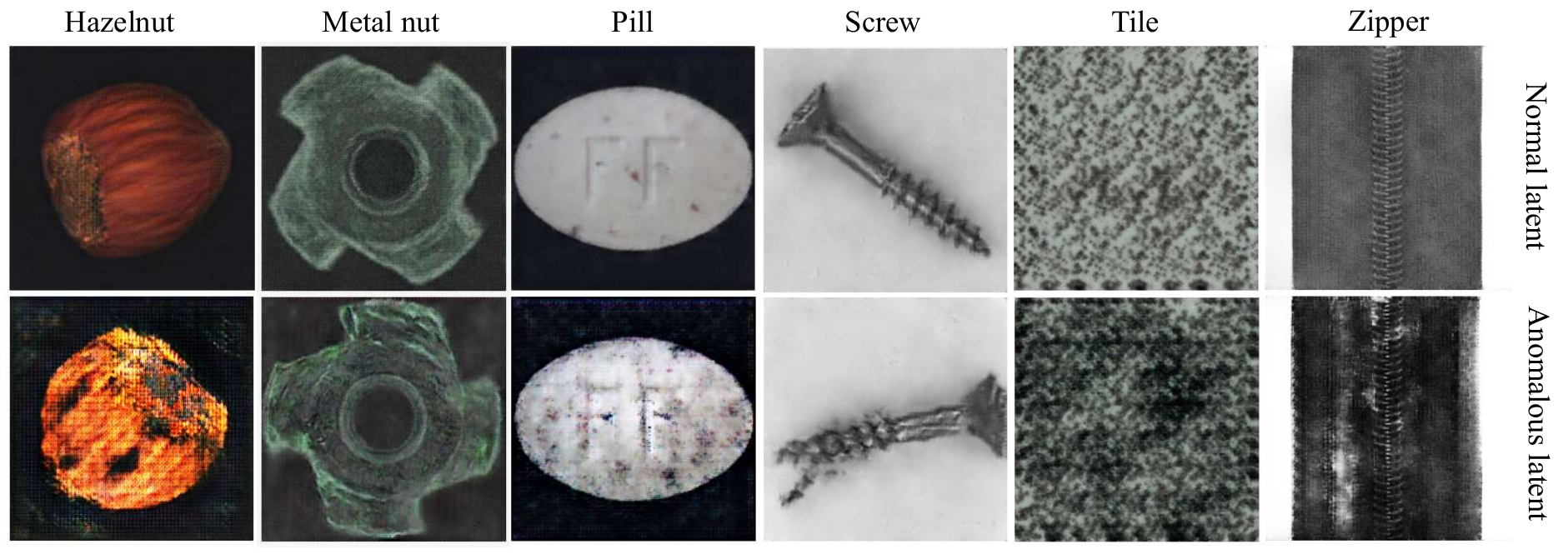}
    \caption{Images generated by ALGAN-image. Top row: Fake-normal data from normal latent variables $N(0,I)$. Bottom row: Fake-anomalous data from anomalous latent variables $N(0,\sigma^{2}I)$, where
    $\sigma=4$.}
    \label{fig:generated_images}
\end{figure*}

This paper discusses anomaly detection using only normal training data. Because preparing a large amount of anomalous data is difficult, training without anomalous data is a preferable approach. In this line of research, a wide variety of methods have already been proposed based on traditional machine learning and statistical techniques, such as one-class classification, likelihood, nearest neighbors, and clustering. See Chandola et al.~\cite{Chandola_etal_2009} for a comprehensive survey of traditional approaches.

Recently, deep learning methods such as representation learning and deep generative models have been successfully applied to anomaly detection without using anomalous data for training.
See~\cite{chalapathy2019deep, ruff2020unifying, pang2021deepanomreview} for a comprehensive survey of the deep learning approaches. Taking advantage of the effective representation of deep learning, the features obtained by a pre-trained model, such as VGG~\cite{Simonyan15} and ResNet~\cite{he2016deep}, are applied to unsupervised anomaly detection~\cite{andrews2016transfer}.  
Because Generative Adversarial Networks (GANs,~\cite{goodfellow2014generative}) can learn a generative model of normal data through a discrimination task, they have been combined for anomaly detection in various ways \cite{schlegl2017unsupervised,zenati2018efficient,akcay2018ganomaly,akccay2019skip,sabokrou2018adversarially}. More details on anomaly detection without using anomalous data for training are presented in Section~\ref{sec:related}.

In this paper, we propose a method that improves anomaly detection performance by generating pseudo-anomalous data from only normal training data using GANs.
Unlike the standard usage of GANs, the generator used in the proposed method provides pseudo-anomalous data as well as fake-normal data, by introducing anomalous states in the latent variable.
We call this model \underline{A}nomalous \underline{L}atent \underline{GAN} (ALGAN).  
Note that the discriminator of a standard GAN is not necessarily suitable for distinguishing between normal and anomalous data. It is trained to discriminate between real and fake data such that in successful learning, the two classes are almost similar.
By contrast, when training is successful, the discriminator of ALGAN distinguishes between the group of real-normal data and the group of fake-normal and pseudo-anomalous data.

Some state-of-the-art anomaly detection methods specialize in product appearance inspection from images.
For example, DifferNet~\cite{RudWan2021} concatenates feature vectors from three different resolution images and uses them to train a model; furthermore, PatchCore~\cite{roth2021towards} uses the hierarchical patches of features to achieve effective and fast performance. However, these methods presume a pre-trained model and cannot be trained directly from the image data. By contrast, the proposed ALGAN can be trained using both images and features.

To apply anomaly detection in real-world applications, it is important to consider computational resources~\cite{chalapathy2019deep}. Normal training data can be collected efficiently; however, training time increases with the amount of training data. In addition, real-time prediction is significant in the real world~\cite{sabokrou2015real}, where the data generation speed has increased~\cite{ahmad2017unsupervised}.
Reducing computational costs will contribute to the expansion of the application~\cite{menghani2021efficient}.
The proposed ALGAN achieves fast training and prediction times.

The contributions of this study are as follows:
\begin{itemize}

\item We propose a novel method for generating pseudo-anomalous data: adding pseudo-anomalous data to GAN training improves the anomaly detection performance of the discriminator.

\item The proposed method can be applied to both images and feature vectors. Experimental results show that it achieves a state-of-the-art performance compared with image-based methods and comparable ability to feature-based methods.

\item The proposed ALGAN achieved a remarkably fast prediction time, 10.4 to 54.6 times faster than other image-based methods on the benchmark MVTec-AD dataset.

\end{itemize}

The remainder of this paper is organized as follows.
We compare our proposed method to relevant anomaly detection methods in Section~\ref{sec:related}.
Section~\ref{sec:method} presents preliminaries on standard GANs, details of the proposed method, and intuition for pseudo-anomalous data.
Section~\ref{sec:experiments} provides the implementation details, datasets, and evaluation method. Section~\ref{sec:results} examines the anomaly detection performances on various datasets and presents the computation time, stability, and ablation study results.
Section~\ref{sec:discussion} discusses the advantages of this study and possible future work. Section~\ref{sec:conclusion} presents concluding remarks.

\section{Related Work}
\label{sec:related}
\subsection{GAN-Based Anomaly Detection}

\begin{figure*}[!t]
    \centering
    \includegraphics[width=\linewidth]{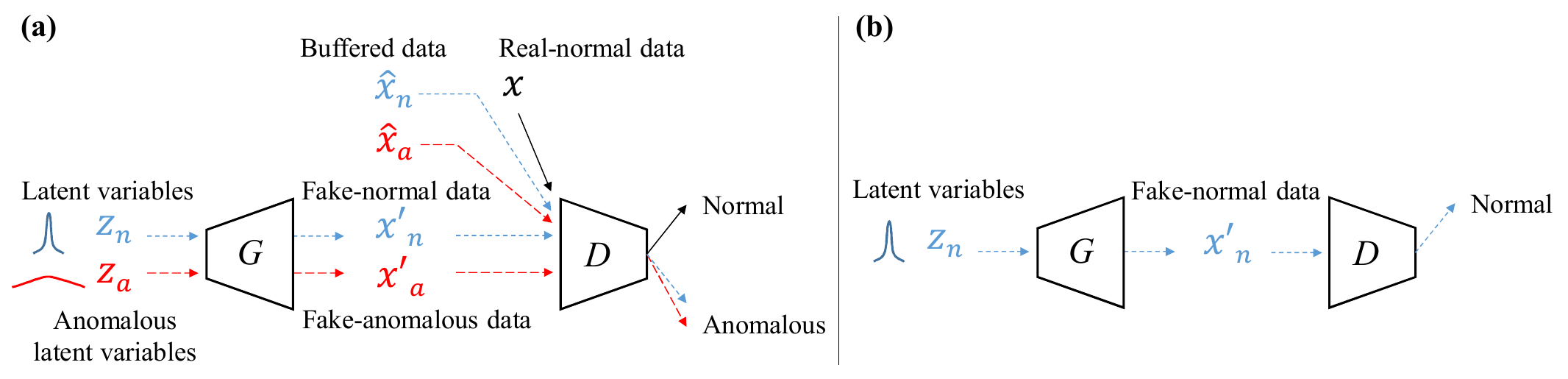}
    \caption{Overview of ALGAN training. (a) Procedure for training the discriminator $D$ with real-normal data $x$ (black), latent variables $z_n \sim N(0, I)$ (blue), anomalous latent variables $z_a \sim N(0, \sigma^2 I)$ (red), fake-normal data $x'_n$ (blue), fake-anomalous data $x'_a$ (red), buffered fake-normal data $\hat{x}_n$ (blue), and buffered fake-anomalous data $\hat{x}_a$ (red). (b) Procedure for training the generator $G$.}
    \label{fig:train_overview}
\end{figure*}

A vast body of literature already exists on generative models for anomaly detection; therefore, we present only representative studies that use GANs. Anomaly detection methods using GANs can be divided into two categories: those using reconstruction errors and those using one-class classifiers. In early studies~\cite{schlegl2017unsupervised, zenati2018efficient}, the latent variable corresponding to given test data was estimated, and the reconstruction error of the image generated from the latent variable was used as the anomaly score.
GANomaly~\cite{akcay2018ganomaly} uses two encoders that estimate the latent variables of the input and generated images and detects anomalies using the reconstruction error between two latent variables.
Skip-GANomaly~\cite{akccay2019skip} is an improvement that evaluates the anomaly score using the features extracted from the middle layer of the discriminator. Because a generator with an encoder--decoder structure trained only on normal data may not properly reconstruct anomalous data, ALOCC~\cite{sabokrou2018adversarially} exploits the one-class classification of the reconstructed images. 
However, these methods fail to detect anomalies when data reconstruction is successful. By contrast, the proposed method does not depend on the reconstruction error and discriminates directly.

\subsection{Anomaly Detection with Pre-Trained Models}
Remarkable performance has been demonstrated in a recent work~\cite{bergman2020deep} in which feature representations  were exploited from pre-trained models on the ImageNet dataset~\cite{deng2009imagenet}. The method in~\cite{rippel2021modeling} extracts the hierarchical features of normal data from a pre-trained EfficientNet~\cite{tan2019efficientnet}, fits them with a multivariate normal distribution, and uses the Mahalanobis distance as the anomaly score. SPADE~\cite{cohen2020sub} extracts hierarchical features, concatenates them, and performs anomaly detection at the image and fine-grained pixel levels using the k-nearest neighbors. PaDiM~\cite{defard2021padim} separates hierarchical features into patches and stores the mean and covariance in a memory bank to measure the Mahalanobis distance. PatchCore~\cite{roth2021towards} subsamples patches of hierarchical features to achieve a high performance and fast prediction.
DifferNet~\cite{RudWan2021} uses a flow-based model~\cite{DBLP:conf/iclr/DinhSB17}, which typically is computation-intensive, and utilizes pre-trained model features~\cite{NIPS2012_c399862d} that reduce data dimensionality and computational costs for likelihood-based anomaly detection.
Most of these methods use traditional anomaly detection techniques that require the use of features, and images cannot be used directly for training. By contrast, the proposed method can use both types of training data.

\subsection{Pseudo-Anomalous Data}
Some methods consider the generation of pseudo-anomalous data. Data immaturely generated during the training process of GANs have been used as pseudo-anomalous data for training~\cite{chatillon2020history, zaheer2020old, pourreza2021g2d}. In CutPaste~\cite{li2021cutpaste}, patches of random sizes and angles are cut out from an image and randomly pasted onto the image. The classifier is trained either from scratch or fine-tuned using normal and pseudo-anomalous data. The feature representation by the classifier is then used to calculate the anomaly score based on the Gaussian density assumption.
The proposed method generates pseudo-anomalous data by introducing anomalous states into latent variables other than using data generated by an immature generator. Thus, it is less biased than techniques that generate pseudo-anomalous data using prior knowledge.

\section{Proposed Method}

\label{sec:method}
\subsection{Overview}
Fig.~\ref{fig:train_overview} illustrates the training procedure of our proposed ALGAN. The generator is trained in the same manner as in standard GANs. Two additional data types are employed to train the discriminator. One of the data types is generated from the anomalous latent variable, and the other is a buffer of data generated during the training process. The buffer size is twice as large as the training data, and in each epoch, a portion of the old buffer is replaced with newly generated data.

\subsection{Generative Adversarial Networks}
\label{sec:GAN}
GANs~\cite{goodfellow2014generative} replace distribution modeling with a discrimination problem: the generator $G(z; \theta)$ maps latent variables $z$ to the data space, and the discriminator $D(x; \phi)$ distinguishes between real data $x$ and the generated samples $x'=G(z)$. 
The discriminator outputs the probability of the input, and real data and generated samples are labeled with probability $1$ and $0$, respectively. The discriminator is trained to maximize the log-likelihoods $\text{log}(D(x))$ and $\text{log}(1-D(G(z)))$. Conversely, the generator is trained to minimize $\text{log}(1-D(G(z)))$ to fool the discriminator. Both network objectives are given by the following equations:
\begin{align}
\label{eq:gan_dis}
\underset{D}{\text{max}} \:
\Bigl(&\mathbb{E}_{x \sim p_{data}(x)}[\text{log}(D(x))]  \nonumber\\
+ &\mathbb{E}_{z \sim p_{z}(z)}[\text{log}(1-D(G(z)))]\Bigr),\\
\label{eq:gan_gen}
\underset{G}{\text{min}} \:
\Bigl(&\mathbb{E}_{z \sim p_{z}(z)}[\text{log}(1-D(G(z)))]\Bigr), 
\end{align}
where $p_{data}$ and $p_{z}$ denote the distributions of real data and latent variables, respectively.

When the discriminator can no longer distinguish between real data and generated samples, the generator approximately realizes a sampler from the real data distribution.
The objective function of the GANs is obtained by combining~\eqref{eq:gan_dis} and~\eqref{eq:gan_gen} as follows:

\begin{align}
\underset{G}{\text{min}} \: \underset{D}{\text{max}} \:
\Bigl(&\mathbb{E}_{x \sim p_{data}(x)}[\text{log}(D(x))]  \nonumber\\
+&\mathbb{E} _{z \sim p_{z}(z)}[\text{log}(1-D(G(z)))]\Bigr).
\label{eq:gan_minmax}
\end{align}

Given the optimal generator, the discriminator can distinguish between two similar classes. Thus, it is nontrivial to determine whether such a discriminator is suitable for identifying real and anomalous data.
When GANs are trained on a dataset $\{x\}$ that includes only normal data, the discriminator has poor discrimination performance on anomalous data~\cite{schlegl2017unsupervised}.
This suggests that the discrimination boundary of GANs is not specialized in the one-class classification of anomaly detection.

\subsection{Pseudo-Anomalous Data}
We introduce two types of pseudo-anomalous data for training. They are in addition to the $x$ and $x'_n=G(z_n)$  used in the standard GANs described in Section~\ref{sec:GAN}, where $x$, $x'_n$, and $z_n$ are defined as real-normal data, fake-normal data, and latent variables from $N(0, I)$, respectively.

\begin{figure*}[t!]
    \centering
    \includegraphics[scale=0.87]{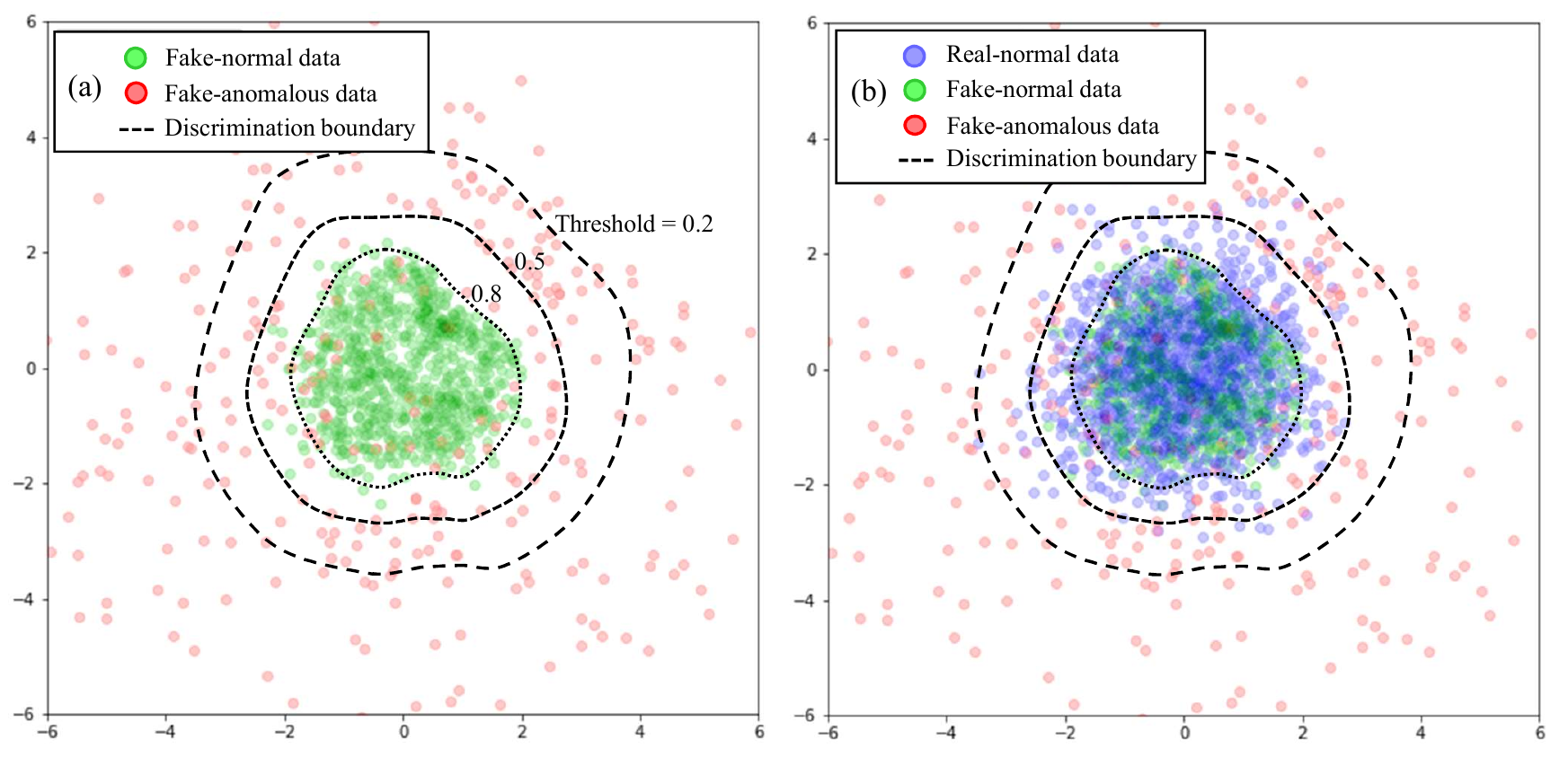}
    \caption{Intuitive interpretation of fake-anomalous data in the toy problem: fake-anomalous data are added to training GANs that map 100-dimensional latent variables to 2-dimensional normal distribution. (a) Fake-anomalous data $x'_{a}$ (red) generated by $\sigma = 4$ are distributed to surround the fake-normal data $x'_{n}$ (green). (b) Real-normal data $x$ (blue) are overlaid on the left figure. From the figure, it can be seen that the generator has been trained successfully and has generated fake-normal data that approximate the real-normal data distribution. Thus, the discriminator cannot distinguish between the real-normal and fake-normal data. By contrast, the fake-anomalous data can be properly distinguished by adjusting the discrimination threshold because they are distributed outside.}
    \label{fig:toy_problem}
\end{figure*}
\begin{algorithm}[t!]
\caption{Training algorithm of ALGAN.}
\label{alg:algorithm}
\textbf{Notation}: Number of batches, $m$; latent variables for $D$, $z_{d}$; latent variables for $G$, $z_{g}$.\\
\textbf{Hyperparameters}: Training epochs ($e$), update frequency of latent variables ($n_{z}$), ratio of normal and anomalous latent variables ($\alpha$), standard deviation of anomalous latent ($\sigma$), and number of updates for $D$ ($n_{dis}$).
\begin{algorithmic}[1] 
\FOR{$i=1, \cdots, e$}
    \IF {$i$ mod $n_z=0$}
        \STATE Sample $z_d \sim \alpha N(0,I)$ and $(1-\alpha)N(0,\sigma^{2}I)$
        \STATE Sample $z_g \sim N(0,I)$
    \ENDIF
    \FOR{$j=1, \cdots, m$}
    \STATE Sample $x \sim p_{data}$
        \FOR{$k=1, \cdots, n_{dis}$}
            \STATE $x' \leftarrow G_{\theta}(z_d)$
            \IF {$i=1$}
                \STATE $Loss_{D} \leftarrow D_{\phi}(x) + D_{\phi}(x')$
            \ELSE
                \STATE Sample buffered data $\hat{x} \sim \text{Buffer}$
                \STATE $Loss_{D} \leftarrow D_{\phi}(x)+D_{\phi}(x')+D_{\phi}(\hat{x})$
            \ENDIF
            \STATE ${\phi} \leftarrow Adam(Loss_{D}, {\phi})$
        \ENDFOR
        \STATE $\text{Buffer} \leftarrow x'$ 
        \STATE $Loss_G \leftarrow D_{\phi}(G_{\theta}(z_g))$
        \STATE ${\theta} \leftarrow Adam(Loss_G, {\theta})$
    \ENDFOR
\ENDFOR
\STATE \textbf{return} $D_{\phi}, G_{\theta}$
\end{algorithmic}
\label{algo:algo_01}
\end{algorithm}

The first type of pseudo-anomalous data is called \emph{fake-anomalous data}.
ALGAN utilizes the anomalous latent variables $z_{a}\sim N(0,\sigma^2 I)$ with a larger variance $(\sigma>1)$ to generate fake-anomalous data $x'_a=G(z_a)$.
See Fig.~\ref{fig:generated_images} for examples of fake-anomalous data.
The fake-anomalous images are slightly degraded compared with the fake-normal images.

The other type of pseudo-anomalous data is called \emph{buffered data} $\hat{x}=\{\hat{x}_{n},\hat{x}_{a}\}$, which are defined as generated samples during the early stage of the training process. These are expected to differ from the real-normal data.
During training, the fake samples $x'=\{x'_{n},x'_{a}\}$ are stored and used as buffered data.

\subsection{Training Methodology}
\label{sec:objectve}


The discriminator is trained to maximize log-likelihoods $\text{log}D(x)$, $\text{log}(1-D(x'))$, and $\text{log}(1-D(\hat{x}))$ to distinguish real-normal data $x$ from other data. Conversely, the generator is trained to generate fake-normal data from $z_n$, and minimize log-likelihood $\text{log}(1-D(G(z_n)))$ to fool the discriminator. Both network objectives are given by the following equations:
\begin{align}
\label{eq:algan_dis}
&\underset{D}{\text{max}} \:
\Bigl(\mathbb{E}_{x}[\text{log}(D(x))] + \xi\mathbb{E}_{z_n}[\text{log}(1-D(x'))] \nonumber\\
&\;\;\;\;\;\;\;\;\;\;\;\;\;\;\;\;\;\;\;\;\;\:+  (1-\xi)\mathbb{E}_{z_n}[\text{log}(1-D(\hat{x}))]\Bigr),\\
\label{eq:algan_gen}
&\underset{G}{\text{min}} \:
\Bigl(\mathbb{E}_{z_n}[\text{log}(1-D(G(z)))]\Bigr), 
\end{align}
where $\xi$ is the ratio of generated data to buffered data. Let the ratio of fake-normal data to fake-anomalous data be $\alpha$; then, the objective of the discriminator is given by the following equation:
\begin{align}
\label{eq:algan_dis_alpha}
 \underset{D}{\text{max}} \; \Bigl(\mathbb{E}_{x}[\text{log}(D(x))] +& \alpha \lbrace \xi \mathbb{E}_{z_n}[\text{log}(1-D(x'_n)] \nonumber\\
+(1-\xi)\mathbb{E}_{\hat{x}_n}&[\text{log}(1-D(\hat{x}_n))] \rbrace \nonumber\\
+(1-\alpha)\lbrace\xi\mathbb{E}_{z_a}\,&[\text{log}(1-D(x'_a))] \nonumber\\
+(1-\xi)\mathbb{E}_{\hat{x}_a}&[\text{log}(1-D(\hat{x}_a))] \rbrace \Bigr).
\end{align}

Thus, the discriminator of ALGAN learns the discrimination boundary between real-normal data and the other types of data. The objective function of ALGAN is obtained by combining~\eqref{eq:algan_gen} and~\eqref{eq:algan_dis_alpha} as follows:
\begin{align} 
\underset{G}{\text{min}}\, 
\underset{D}{\text{max}}\, \Bigl(\mathbb{E}_{x}[\text{log}(D(x))] +& \alpha \lbrace \xi \mathbb{E}_{z_n}[\text{log}(1-D(G(z_n)))] \nonumber\\
+(1-\xi)\mathbb{E}_{\hat{x}_n}&[\text{log}(1-D(\hat{x}_n))] \rbrace \nonumber\\
+ (1-\alpha) \lbrace \xi \mathbb{E}_{z_a}\,&[\text{log}(1-D(x'_a))] \nonumber\\
+ (1-\xi) \mathbb{E}_{\hat{x}_a}&[\text{log}(1-D(\hat{x}_a))] \rbrace \Bigr).
\label{eq:ours}
\end{align}
The proposed method follows an adversarial training procedure (Fig.~\ref{fig:train_overview}). It provides a discrimination boundary not only for the real-normal and fake-normal data but also for the real-normal and pseudo-anomalous data, the latter of which has a broader support of the distribution. 
As the training progresses, the generator produces samples that resemble real-normal data, and the discriminator cannot distinguish between real-normal and fake-normal data. The pseudo-anomalous data are clearly different from the real-normal data; therefore, the discrimination boundary of the discriminator is used to classify them.

The pseudo-code for training is presented in Algorithm~\ref{alg:algorithm}.
A feature of ALGAN training is the method for updating the parameter $\phi$ of the discriminator.
In line 3 of the pseudo-code, normal and anomalous latent variables are sampled, and in line 9, fake-normal and fake-anomalous data are produced from the generator.
In line 11, loss is calculated by identifying both types of data, and in line 16, the parameter $\phi$ of the discriminator is updated.
Fake-normal and fake-anomalous data are buffered in line 18.
As there are no buffered data in the first epoch, the conditional branch for $i=1$ is required in line 10. The loss of buffered data is calculated in addition to fake-normal and fake-anomalous data in line 14 after the branch. The parameter $\theta$ of the generator is updated by only fake-normal data from normal latent variables in lines 4, 19, and 20.

Fig.~\ref{fig:toy_problem} and the accompanying movie provide an intuitive understanding of the training for anomaly detection using fake-anomalous data. 
The generator is trained to produce fake-normal data that approximate the distribution of real-normal data. Given anomalous latent variables, the generator produces fake-anomalous data with a large variance. The discriminator is trained to distinguish between real-normal data and the other types of data. Even after real-normal and fake-normal data become indistinguishable, fake-anomalous data can be identified because the discrimination boundary is laid between real-normal and fake-anomalous data.

\subsection{Anomaly Detection}
In ALGAN, the real and fake labels are assigned probability  $1$ and $0$, respectively. The real label corresponds to normal data; thus, the anomaly detection rule is given by the following:
\begin{equation}
\text{ALGAN}(x) =
\begin{cases}
    \text{Normal,} & \text{if $D(x) >$ threshold,} \\
    \text{Anomalous,} & \text{otherwise}.
\end{cases}
\end{equation}
Because the significance of false positives or false negatives depends on their application, the threshold is chosen by the user.

\section{Experiments}
\label{sec:experiments}

\begin{figure*}[!t]
    \includegraphics[width=\linewidth]{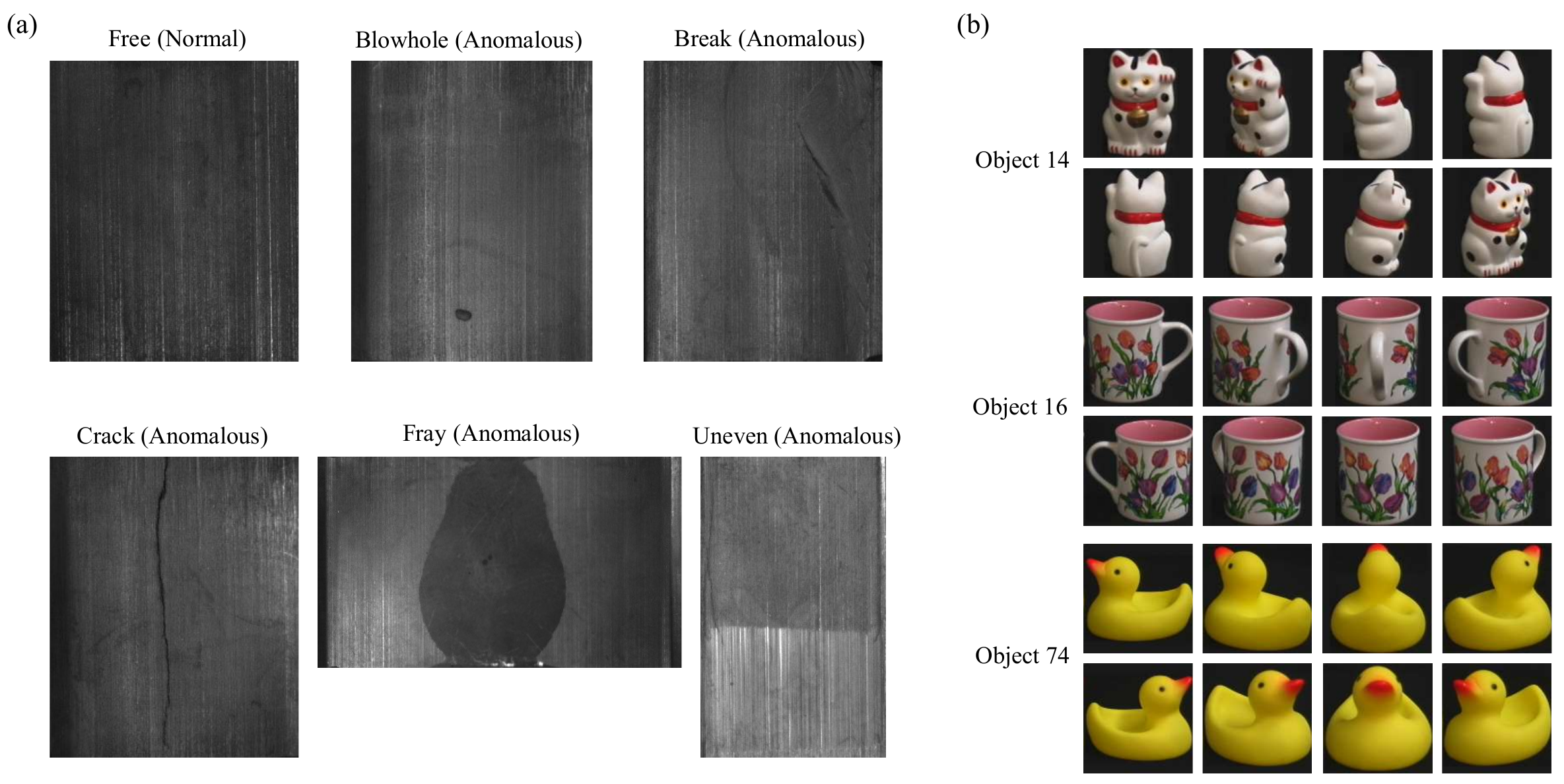}
    \caption{Overview of the experimental datasets. (a) Magnetic Tile Defects dataset (MTD); (b) Columbia University Image Library dataset (COIL-100)}
    \label{fig:MTD_COIL}
\end{figure*}

An advantage of ALGAN is that it can be used for both the images and features extracted from a pre-trained model, whereas some state-of-the-art methods for visual inspection depend on features~\cite{cohen2020sub, defard2021padim, roth2021towards, RudWan2021}. To demonstrate this advantage experimentally, we used two different types of implementations. We call them ALGAN-image and ALGAN-feature, and compare them with the relevant methods.

\subsection{Implementation Details}

\subsubsection{Network Architecture and Hyperparameter}

ALGAN-image employs an architecture similar to that of DCGAN~\cite{DBLP:journals/corr/RadfordMC15}. The generator and discriminator use seven transposed convolutional and convolutional layers, respectively.

For ALGAN-feature, WideResNet101~\cite{zagoruyko2016wide} is applied to the feature extractor to obtain 2048-dimensional vectors from the last block with global average pooling. Both the generator and discriminator have three fully connected layers.

In both architectures, the generator uses batch normalization~\cite{ioffe2015batch} and the ReLU activation function, whereas batch normalization is removed from the output layer. The discriminator employs spectral normalization~\cite{miyato2018spectral} and the Leaky-ReLU activation function.

The networks were optimized using Adam~\cite{kingma:adam} with momentum ${\beta}_{1} = 0$ and ${\beta}_{2} = 0.9$, and the learning rates of the generator and discriminator were set to $2\times10^{-4}$ and $1\times10^{-4}$, respectively. The latent variable $z$ had 100 dimensions, and the standard deviation of the anomalous latent variable used $\sigma=4$. The parameters that determine the ratio of the pseudo-anomalous data were set to $\alpha=0.75$ and $\xi=0.75$. The parameters of Algorithm~\ref{alg:algorithm} were set as $n_{z} = 2$ and $n_{dis} =2$.  The buffer holds twice the batch size, half of which is randomly replaced by newly generated data.

The comparison methods were implemented using a DCGAN-like architecture similar to ALGAN-image. In GANomaly~\cite{akcay2018ganomaly} and Skip-GANomaly~\cite{akccay2019skip}, the dimensions of the latent variables were set to 100 and 512, respectively. The image resolution used for each method was $256 \times 256$. All the models were trained using a batch size of 16.

\subsubsection{Software and Hardware}
\label{sec:hardware}
All the models were implemented using Python 3.8.8 and PyTorch 1.8.1 on Ubuntu 20.04 LTS. An AMD EPYC 7542 32-core processor with 512 GB memory and an NVIDIA A100-SXM4 40 GB GPU were used for the computations.

\subsection{Datasets}
For performance evaluation, we used three different datasets.

\subsubsection{MVTec-AD Dataset}
This dataset, designed for visual inspection, consists of 5,354 images, comprising five texture and 10 object categories~\cite{bergmann2019mvtec}. The training set contains 3,629 defect-free (normal) images, and the test set contains 467 defect-free (normal) images and 1,258 defective (anomalous) images.

\subsubsection{Magnetic Tile Defects Dataset (MTD)}
This dataset consists of grayscale images with different aspect ratios, including 952 defect-free (normal) images and 392 images containing five defect types (anomalous)~\cite{huang2020surface} (Fig.~\ref{fig:MTD_COIL}).

\subsubsection{COIL-100 Dataset}
This dataset contains 100 different object categories and 7,200 images~\cite{nene1996columbia}. Each object category has 72 images rotated every $5^\circ$ (Fig.~\ref{fig:MTD_COIL}).

\subsubsection{Splitting into Training and Validation Sets}
Because the training sets of MVTec-AD and MTD contain no defective (anomalous) images, 50\% of their respective test sets was used as the validation set, which was used as the stopping rule.

In MTD, the test set is not provided separately from the training data;
therefore, 50\%, 25\%, and 25\% of the defect-free (normal) images were used for training, validation, and testing, respectively.

For COIL-100, 10 categories were selected for normal data and the remaining 90 categories for anomalous data. We used 60\%, 20\%, and 20\% of the normal data for training, validation, and testing, respectively, and 50\% of the anomalous data were used for validation and 50\% for testing.

\subsubsection{Pre-Processing and Data Augmentation}

In MVTec-AD using ALGAN-image, images were resized to $256 \times 256$. To highlight defects, two images were concatenated with the original image and used as an input image: one was created with max pooling and the other with average pooling in the channel axis dimension. For the texture categories, vertical and horizontal flips were applied (however, only horizontal flip was applied for wood). In the object categories, vertical flip, horizontal flip, and random rotation were applied to Bottle and Hazelnut. Horizontal flip was applied to Toothbrush, Transistor, and Zipper. Toothbrush was converted into grayscale. Random rotation was applied to Metal nut and Screw. Cable, Capsule, and Pill were only resized.

In MVTec-AD using ALGAN-feature, the texture categories were resized to $224 \times 224$. Furthermore, the object categories were resized to $256 \times 256$, and then center-cropped to $224 \times 224$. Following Rudolph et al.'s procedure~\cite{RudWan2021}, we applied 24 and 64 different angular rotations during training and prediction, respectively, but using a single-resolution image.

The images in MTD and COIL-100 were resized to $256 \times 256$ for ALGAN-image and $224 \times 224$ for ALGAN-feature. For ALGAN-image, a horizontal flip was applied, and for ALGAN-feature, Rudolph et al.'s procedure~\cite{RudWan2021} was applied for both training and prediction.

\subsection{Evaluation Method and Metric}
\label{sec:evalandmetric}
Accounting for the randomness of the dataset split, we performed 10 experiments with different seed values for each dataset.
Performance was evaluated in terms of the area under the receiver operating characteristic (AUROC) curve, which is obtained by moving the classification threshold for the ratio of true-positive and false-positive rates. 
ALGAN-image was trained on 512 epochs and ALGAN-feature on 192 epochs. The performances were validated every eight epochs on the validation set, and the model that showed the best AUROC was saved. The best model was evaluated on the test set after training.

\section{Results}
\label{sec:results}

\begin{table}[!t]
\small
\begin{center}
\caption{Results obtained on MVTec-AD. ALGAN-image is compared with methods trained on image data. Top row: mean AUROC. Bottom row: standard deviation. We report the results of 10 experiments using each method. The best performance for each category is indicated in boldface.}
\begin{tabular}{@{}lrrrr@{}}
\toprule
 & \multicolumn{1}{c}{GANomaly} & \multicolumn{1}{c}{\begin{tabular}[c]{@{}c@{}}Skip-\\ GANomaly\end{tabular}} & \multicolumn{1}{c}{ALOCC} & \multicolumn{1}{c}{\begin{tabular}[c]{@{}c@{}}ALGAN\\ -image\end{tabular}} \\ \midrule
\multirow{2}{*}{Carpet} & 0.803 & 0.829 & 0.736 & \textbf{0.846} \\
 & 0.070 & 0.056 & 0.054 & 0.041 \\ \midrule
\multirow{2}{*}{Grid} & 0.924 & 0.816 & 0.847 & \textbf{0.938} \\
 & 0.088 & 0.079 & 0.087 & 0.057 \\ \midrule
\multirow{2}{*}{Leather} & 0.796 & 0.787 & 0.768 & \textbf{0.920} \\
 & 0.079 & 0.095 & 0.042 & 0.041 \\ \midrule
\multirow{2}{*}{Tile} & 0.852 & 0.941 & 0.648 & \textbf{0.914} \\
 & 0.034 & 0.041 & 0.071 & 0.034 \\ \midrule
\multirow{2}{*}{Wood} & 0.939 & 0.969 & 0.849 & \textbf{0.972} \\
 & 0.029 & 0.020 & 0.062 & 0.022 \\ \midrule
\multirow{2}{*}{Bottle} & 0.704 & 0.674 & 0.801 & \textbf{0.948} \\
 & 0.063 & 0.096 & 0.075 & 0.024 \\ \midrule
\multirow{2}{*}{Cable} & 0.686 & 0.636 & 0.686 & \textbf{0.877} \\
 & 0.057 & 0.064 & 0.053 & 0.036 \\ \midrule
\multirow{2}{*}{Capsule} & 0.717 & 0.691 & 0.707 & \textbf{0.805} \\
 & 0.075 & 0.068 & 0.073 & 0.083 \\ \midrule
\multirow{2}{*}{Hazelnut} & 0.765 & \textbf{0.955} & 0.697 & 0.836 \\
 & 0.058 & 0.036 & 0.111 & 0.047 \\ \midrule
\multirow{2}{*}{Metal nut} & 0.698 & 0.553 & 0.771 & \textbf{0.811} \\
 & 0.064 & 0.105 & 0.066 & 0.060 \\ \midrule
\multirow{2}{*}{Pill} & 0.772 & 0.787 & 0.659 & \textbf{0.819} \\
 & 0.050 & 0.083 & 0.043 & 0.053 \\ \midrule
\multirow{2}{*}{Screw} & 0.567 & \textbf{1.000} & 0.938 & 0.811 \\
 & 0.357 & 0.001 & 0.076 & 0.090 \\ \midrule
\multirow{2}{*}{Toothbrush} & 0.774 & 0.808 & 0.778 & \textbf{0.933} \\
 & 0.098 & 0.084 & 0.102 & 0.048 \\ \midrule
\multirow{2}{*}{Transistor} & 0.783 & 0.769 & 0.745 & \textbf{0.865} \\
 & 0.067 & 0.094 & 0.085 & 0.074 \\ \midrule
\multirow{2}{*}{Zipper} & 0.693 & 0.675 & 0.656 & \textbf{0.879} \\
 & 0.052 & 0.070 & 0.079 & 0.025 \\ \midrule
mean & 0.765 & 0.793 & 0.752 & \textbf{0.878} \\ \bottomrule
\end{tabular}
\label{table:mvtecimage}
\end{center}
\end{table}

\begin{table*}[]
\small
\begin{center}
\caption{Results obtained on MVTec-AD. ALGAN-feature is compared with methods learned from features of pre-trained models. For DifferNet and ALGAN-feature, we report the mean AUROC results of 10 experiments. The best performance for each category is indicated in boldface.}
\begin{tabular}{@{}lcrrrrrrr@{}}
\toprule
 & SPADE & \multicolumn{1}{c}{Mah.AD} & \multicolumn{1}{c}{PaDiM} & \multicolumn{1}{c}{CutPaste} & \multicolumn{1}{c}{PatchCore-25} & \multicolumn{1}{c}{GANomaly} & \multicolumn{1}{c}{DifferNet} & \multicolumn{1}{c}{ALGAN-feature} \\ \midrule
Texture & - & 0.978 & 0.990 & 0.962 & \textbf{0.991} & 0.775 & 0.915 & 0.925 \\
Object & - & 0.950 & 0.972 & 0.955 & \textbf{0.992} & 0.755 & 0.934 & 0.905 \\
ALL & \multicolumn{1}{r}{0.855} & 0.958 & 0.979 & 0.961 & \textbf{0.991} & 0.761 & 0.927 & 0.911 \\ \bottomrule
\end{tabular}
\label{table:mvtecfeat}
\end{center}
\end{table*}

\subsection{Anomaly Detection on MVTec-AD}
\subsubsection{Training with Image Data}
The results on the test data with the model with the best AUROC for validation (Section~\ref{sec:evalandmetric}) are listed in Table~\ref{table:mvtecimage}. ALGAN-image significantly outperformed other state-of-the-art image-based methods, such as GANomaly, Skip-GANomaly, and ALOCC. ALGAN-image showed an average accuracy of more than 10\% compared with the others and attained the best accuracy for 13 out of the 15 categories.

Our method uses the discriminator to distinguish between normal and anomalous directly from the images.
In the Hazelnut and Screw categories, the rotation angle of the object is different in each image.
Thus, small changes in image details caused by slight defects may be buried by large changes in the image caused by rotation.
Consequently, the anomalous data detection performance could not be better than that of the comparison methods in these categories.

Because the comparison methods use reconstruction error, they fail to detect anomalous data if the reconstruction is successful. By contrast, our proposed method, ALGAN-image, uses the discriminator to classify the data and does not suffer from detection errors due to reconstruction.

\subsubsection{Training with Pre-Trained Features}
The results for the feature-based methods are listed in Table~\ref{table:mvtecfeat}. The results for DifferNet are the means of 10 different test datasets (see Section~\ref{sec:evalandmetric}). The results for GANomaly were obtained from~\cite{RudWan2021} and the other methods from their respective papers~\cite{cohen2020sub, rippel2021modeling, defard2021padim, li2021cutpaste, roth2021towards}. ALGAN-feature achieved results comparable to those of DifferNet, whereas PatchCore-25, PaDiM, and CutPaste achieved better performance. Note, however, that the PatchCore, PaDiM, and CutPaste methods are strongly specialized for pre-trained features.

ALGAN-feature and DifferNet use only a single feature from the last layer of the pre-trained model; better performing methods use features from multiple layers. 
Using features from multiple layers increases the computational costs owing to their high dimensionality.
For example, PatchCore~\cite{roth2021towards} divides high-dimensional features into patches and subsamples them to select useful patches, which results in much higher computational costs. The prediction time is also longer than that of ALGAN-feature (see Section~\ref{sec:pred_train_times}).
Furthermore, DifferNet~\cite{RudWan2021} concatenates features from three different image resolutions, whereas our proposed method, ALGAN-feature, achieves comparable performance with only one resolution.

\subsection{Anomaly Detection on Other Datasets}

\begin{table*}[]
\small
\begin{center}
\caption{Results obtained on Magnetic Tile Defects. For training with the image data, we report the mean AUROC results of 10 experiments. Numbers in boldface indicate the best performance.}
\begin{tabular}{@{}lccc|cccc@{}}
\toprule
 & \multicolumn{3}{c|}{Training with image data} & \multicolumn{4}{c}{Training with pre-trained features} \\ \midrule
 & GANomaly & Skip-GANomaly & ALGAN-image & PatchCore-10 & GANomaly-feat. & DifferNet & ALGAN-feature \\
AUROC & \multicolumn{1}{r}{0.683} & \multicolumn{1}{r}{0.504} & \multicolumn{1}{r|}{\textbf{0.956}} & \multicolumn{1}{r}{\textbf{0.979}} & \multicolumn{1}{r}{0.766} & \multicolumn{1}{r}{0.977} & \multicolumn{1}{r}{0.824} \\ \bottomrule
\end{tabular}
\label{table:resultsmtd}
\end{center}
\end{table*}

\begin{table}[]
\begin{center}
\caption{Results obtained on COIL-100. Top row: mean AUROC. Bottom row: standard deviation. We report the results of 10 experiments using each method. The best performance is indicated in boldface.}
\begin{tabular}{@{}lrrrr@{}}
\toprule
\multicolumn{1}{l}{} & DifferNet & PatchCore-1 & \begin{tabular}[c]{@{}l@{}}ALGAN\\ -image\end{tabular} & \begin{tabular}[c]{@{}l@{}}ALGAN\\ -feature\end{tabular} \\ \midrule
\multirow{2}{*}{AUROC} & 0.999 & \textbf{1.000} & 0.999 & \textbf{1.000} \\
 & 0.003 & 0.000 & 0.001 & 0.000 \\ \bottomrule
\end{tabular}
\label{table:resultscoil100}
\end{center}
\end{table}

\begin{figure}[t]
    \centering
    \includegraphics[width=\linewidth]{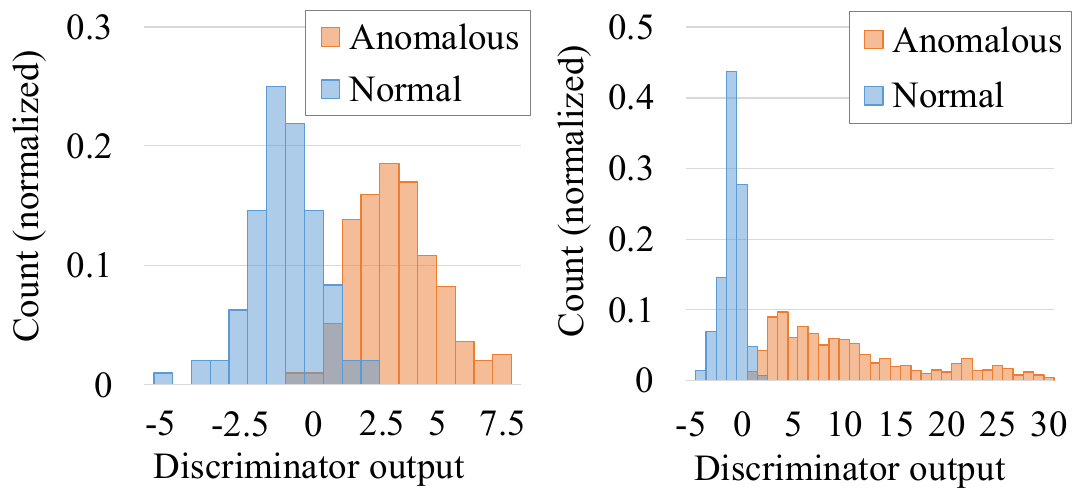}
    \caption{Histograms of raw output values of the discriminator before input to the sigmoid function in ALGAN-image. Left: Magnetic Tile Defects. Right: COIL-100. The sign is reversed so that the horizontal axis represents the anomaly score.}
    \label{fig:histgram}
\end{figure}

The results obtained on MTD and COIL-100 are listed in Tables~\ref{table:resultsmtd} and \ref{table:resultscoil100}, respectively. These are the means of 10 trials with the same hyperparameters as those used for MVTec-AD. ALGAN-image achieved state-of-the-art performance on MTD after training with image data and achieved comparable results to that of PatchCore. On MTD, ALGAN-feature performed worse than ALGAN-image. This may be because the features useful for anomaly detection could not be extracted from the last block of WideResNet-101. Because the features from the deep block in ResNet are biased towards ImageNet~\cite{roth2021towards}, the features from the shallow block should be used to identify the more abstract features required for MTD. 
On the COIL-100 benchmark, all the methods performed almost perfectly.

Fig.~\ref{fig:histgram} shows histograms of the raw output values of the discriminator before they are input into the sigmoid function of ALGAN-image. The distributions of normal and anomalous data are significantly separated. In COIL-100, normal data have a peaky distribution with fewer variations, whereas the distribution of anomalous data exhibits a long tail, reflecting the large variations of anomalous data.

\subsection{Prediction and Training Times}
\label{sec:pred_train_times}

\begin{table*}[]
\begin{center}
\caption{Mean prediction times per single test image for all categories on MVTec-AD. Boldface indicates the best result.}
\begin{tabular}{@{}lccccccc@{}}
\toprule
 & GANomaly & \begin{tabular}[c]{@{}c@{}}Skip-\\      GANomaly\end{tabular} & ALOCC & DifferNet &  \begin{tabular}[c]{@{}c@{}}PatchCore\\      -25\end{tabular}  & \begin{tabular}[c]{@{}c@{}}ALGAN\\      -feature\end{tabular} & \begin{tabular}[c]{@{}c@{}}ALGAN\\      -image\end{tabular} \\ \midrule
milliseconds/image & \multicolumn{1}{r}{109.8} & \multicolumn{1}{r}{126.3} & \multicolumn{1}{r}{577.8} & \multicolumn{1}{r}{489.5} & \multicolumn{1}{r}{278.6} & \multicolumn{1}{r}{218.9} & \multicolumn{1}{r}{\textbf{10.6}} \\ \bottomrule
\end{tabular}
\label{table:predtime}
\end{center}
\end{table*}

\begin{table*}[]
\begin{center}
\caption{Mean training times for all categories on MVTec-AD. Boldface indicates the best result.}
\begin{tabular}{@{}lrrrrcrr@{}}
\toprule
 & \multicolumn{1}{c}{GANomaly} & \multicolumn{1}{c}{\begin{tabular}[c]{@{}c@{}}Skip-\\ GANomaly\end{tabular}} & \multicolumn{1}{c}{ALOCC} & \multicolumn{1}{c}{DifferNet} & \begin{tabular}[c]{@{}c@{}}PatchCore\\ -25\end{tabular} & \multicolumn{1}{c}{\begin{tabular}[c]{@{}c@{}}ALGAN\\ -feature\end{tabular}} & \multicolumn{1}{c}{\begin{tabular}[c]{@{}c@{}}ALGAN\\ -image\end{tabular}} \\ \midrule
Training Epochs & 512 & 512 & 512 & 192 & - & 192 & 512 \\
Training Time (min) & 23 & 47 & 401 & 55 & 98 & 41 & \textbf{22} \\ \bottomrule
\end{tabular}
\label{table:traintime}
\end{center}
\end{table*}

\begin{figure}[t]
    \centering
    \includegraphics[width=\linewidth]{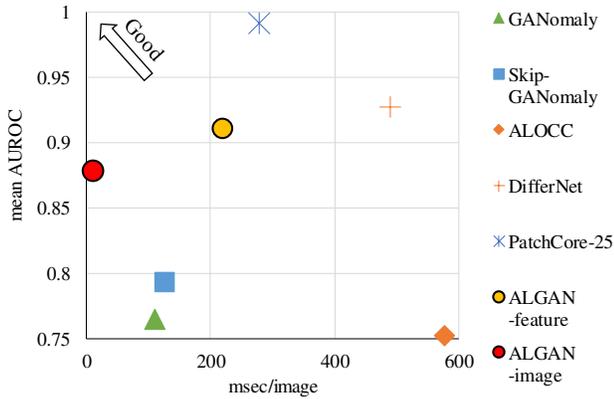}
    \caption{Mean AUROC vs. prediction time for each method on MVTec-AD.}
    \label{fig:auc_vs_msec}
\end{figure}

Table~\ref{table:predtime} compares the prediction times of the models for MVTec-AD. 
We can see that ALGAN-image achieved a significantly faster prediction time (10.6 ms), which is 10.4 to 54.6 times faster than those of the other image-based methods. ALGAN-feature is 1.3 to 2.2 times faster than the other methods trained on the feature. Fig.~\ref{fig:auc_vs_msec} depicts the prediction time and AUROC performance of the selected methods. ALGAN-image is the fastest and has a high AUROC, but it is not the highest. ALGAN-feature is faster than the other feature-based methods while maintaining a competitive AUROC. 
ALGAN and PatchCore exhibit a trade-off between performance and speed.
Considering the fast prediction of ALGAN-image, it can be applied to expensive tasks such as real-time prediction with a large number of bounding boxes obtained from object detection~\cite{liu2020deepobjectdetection}.

Table~\ref{table:traintime} compares the training times obtained on MVTec-AD. ALGAN-image is the fastest among the compared methods, with the default number of epochs described in Section~\ref{sec:evalandmetric}.
Because there is no official implementation of PatchCore, 
we did not use the Faiss library~\cite{johnson2019billionfaiss}\footnote{Our implementation uses the scikit-learn~\cite{scikit-learn} library for core-set selection~\cite{sener2018activekcentergreedy} and random projection~\cite{sinha2020smallrandomprojection}. We confirmed that our implementation produces an AUROC similar to the results in~\cite{roth2021towards}.}, which is used in the original study, but applied our own implementation.

\subsection{Training Stability}

\begin{figure}[t]
    \centering
    \includegraphics[width=\linewidth]{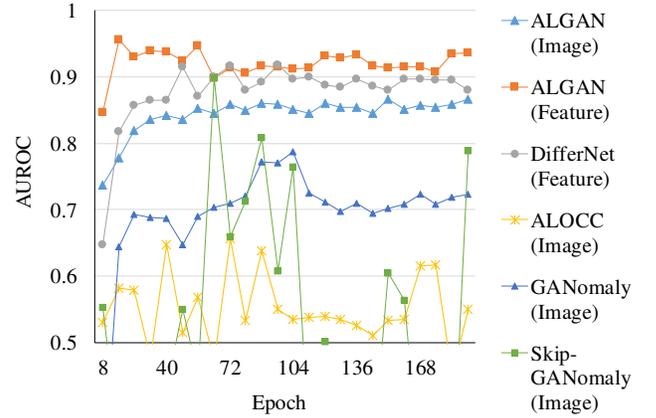}
    \caption{Validation results plotted every 8 epochs during the training of the transistor category. ALGAN-image and ALGAN-feature exhibit stable AUROCs compared with other GAN-based methods.}
    \label{fig:stability}
\end{figure}

Zaheer et al. reported that GAN-based anomaly detection exhibited unstable validation results during the training process~\cite{zaheer2020old}. Upon validation, our proposed method exhibited stable results in terms of AUROC (Fig.~\ref{fig:stability}).

\subsection{Ablation Study}

\begin{table}[!t]
\begin{center}
\caption{Ablation study results obtained on MVTec-AD with ALGAN-image. Each experiment was conducted 10 times. The left side of the mean AUROC column lists the mean and the right side lists the standard deviation. The best performance is indicated in boldface.}
\begin{tabular}{@{}ccrr@{}}
\toprule
Buffered Data & Anomalous Latent & \multicolumn{2}{c}{Mean AUROC} \\ \midrule
 &  & 0.643 & 0.097 \\
\checkmark &  & 0.761 & 0.102 \\
 & \checkmark & 0.779 & 0.104 \\
\checkmark & \checkmark & \textbf{0.878} & 0.049 \\ \bottomrule
\end{tabular}
\label{table:ablation}
\end{center}
\end{table}

Ablation studies were performed to verify the effect of the two types of pseudo-anomalous data on performance. In ALGAN, buffered data and the data generated by anomalous latent variables were used as pseudo-anomalous data. The checkmark in Table~\ref{table:ablation} indicates the type(s) used. Although either of the two can improve performance, using both works best and reduces variance. These results validate that the proposed pseudo-anomalous data are useful for improving anomaly detection performance.

\subsection{Hyperparameter Study}

\begin{table}[!t]
\begin{center}
\caption{Impact of $\sigma$ changes on performance. Top row: mean AUROC. Bottom row: standard deviation. Boldface indicates the best result.}
\begin{tabular}{@{}lrrrrrr@{}}
\toprule
\multicolumn{1}{c}{$\sigma$} & \multicolumn{1}{c}{2} & \multicolumn{1}{c}{3} & \multicolumn{1}{c}{4} & \multicolumn{1}{c}{5} & \multicolumn{1}{c}{6} & \multicolumn{1}{c}{8} \\ \midrule
\multirow{2}{*}{AUROC} & 0.859 & 0.932 & \textbf{0.948} & 0.947 & 0.937 & 0.917 \\
 & 0.058 & 0.042 & 0.024 & 0.025 & 0.033 & 0.036 \\ \bottomrule
\end{tabular}
\label{table:sigma}
\end{center}
\end{table}

\begin{table}[!t]
\begin{center}
\caption{Impact of $\alpha$ and $\xi$ change on performance. First row: mean AUROC. Second row: standard deviation. Boldface indicates the best result.}
\begin{tabular}{@{}llrrrc@{}}
\toprule
\multicolumn{2}{l}{\multirow{2}{*}{}} & \multicolumn{4}{c}{$\xi$} \\
\multicolumn{2}{l}{} & \multicolumn{1}{c}{0.25} & \multicolumn{1}{c}{0.5} & \multicolumn{1}{c}{0.75} & 0.85 \\ \midrule
\multirow{8}{*}{$\alpha$} & \multirow{2}{*}{0.25} & 0.808 & 0.720 & 0.749 & 0.656 \\
 &  & 0.085 & 0.077 & 0.155 & 0.162 \\ \cmidrule(l){2-6} 
 & \multirow{2}{*}{0.5} & 0.755 & 0.814 & 0.892 & 0.794 \\
 &  & 0.148 & 0.081 & 0.068 & 0.157 \\ \cmidrule(l){2-6} 
 & \multirow{2}{*}{0.75} & 0.826 & 0.860 & \textbf{0.948} & 0.808 \\
 &  & 0.138 & 0.048 & 0.024 & 0.122 \\ \cmidrule(l){2-6} 
 & \multirow{2}{*}{0.85} & 0.816 & 0.846 & 0.910 & 0.765 \\
 &  & 0.094 & 0.110 & 0.060 & 0.153 \\ \bottomrule
\end{tabular}
\label{table:alpha_xi}
\end{center}
\end{table}

\begin{table}[!t]
\begin{center}
\caption{Impact of $n_z$ changes on performance. Top row: mean AUROC. Bottom row: standard deviation. Boldface indicates the best result.}
\begin{tabular}{@{}lrrrr@{}}
\toprule
\multicolumn{1}{c}{$n_z$} & \multicolumn{1}{c}{1} & \multicolumn{1}{c}{2} & \multicolumn{1}{c}{3} & \multicolumn{1}{c}{4} \\ \midrule
\multirow{2}{*}{AUROC} & 0.813 & \textbf{0.948} & 0.903 & 0.874 \\
 & 0.121 & 0.024 & 0.035 & 0.104 \\ \bottomrule
\end{tabular}
\label{table:n_z}
\end{center}
\end{table}

\begin{table}[!t]
\begin{center}
\caption{Impact of $n_{dis}$ changes on performance. Top row: mean AUROC. Bottom row: standard deviation. Boldface indicates the best result.}
\begin{tabular}{@{}lrrrr@{}}
\toprule
\multicolumn{1}{c}{$n_{dis}$} & \multicolumn{1}{c}{1} & \multicolumn{1}{c}{2} & \multicolumn{1}{c}{3} & \multicolumn{1}{c}{4} \\ \midrule
\multirow{2}{*}{AUROC} & 0.887 & \textbf{0.948} & 0.943 & 0.933 \\
 & 0.082 & 0.024 & 0.031 & 0.049 \\ \bottomrule
\end{tabular}
\label{table:n_dis}
\end{center}
\end{table}

We also studied the impact of hyperparameters on anomaly detection performance. ALGAN employs the following hyperparameters: $\sigma$ is the standard deviation of the anomalous latent variable, $n_z$ is the epoch frequency to update the latent variables, $n_{dis}$ is the number of updates per batch on the discriminator, and $\alpha$ and $\xi$ are the fake-anomalous and buffered data balanced parameter, respectively. Each experiment was conducted 10 times using the Bottle category of the MVTec-AD dataset.

\subsubsection{Standard Deviation $\sigma$ for Anomalous Latent Variable}
For a small $\sigma$, performance is low and peak performance is reached at $\sigma=4$ or $5$. If the $\sigma$ value is too large, the support for normal and anomalous data may be separated, which can degrade performance (Table~\ref{table:sigma}).

\subsubsection{Balance Parameter $\alpha$ and $\xi$ for Fake-normal and Buffered Data}
When $\alpha$ is small, the effect of fake-anomalous is large, and when $\xi$ is small, the effect of buffered data is large. The performance is high around the values of $\alpha=0.75$, $\xi=0.75$ reported in this study. Recalling~\eqref{eq:algan_dis_alpha}, $\xi$ is multiplied by $\alpha$. When $\alpha=0.25$, $\xi=0.25$, the effect of the fake-anomalous buffer is greater, and the performance is improved. By contrast, when $\alpha=0.85$, $\xi=0.85$, the effect of fake-anomalous and buffered data is smaller, and the performance is lower (Table~\ref{table:alpha_xi}).
The results suggest that the effect of fake-anomalous buffer is large, and the parameters $\alpha$ and $\xi$ should be adjusted so that the effect is not too small.

\subsubsection{Update Frequency for Latent Variable and Discriminator}
In this study, $n_z$ peaked at 2, and $n_{dis}$ was good above 2, but performance decreased slightly when discriminator updates became excessive (Table~\ref{table:n_z},~\ref{table:n_dis}).
Therefore, $n_z$ has a considerable impact on the performance and should be carefully considered and chosen in practical applications.

\subsection{Robustness to other anomalous data}
\begin{figure}[t]
    \centering
    \includegraphics[width=\linewidth]{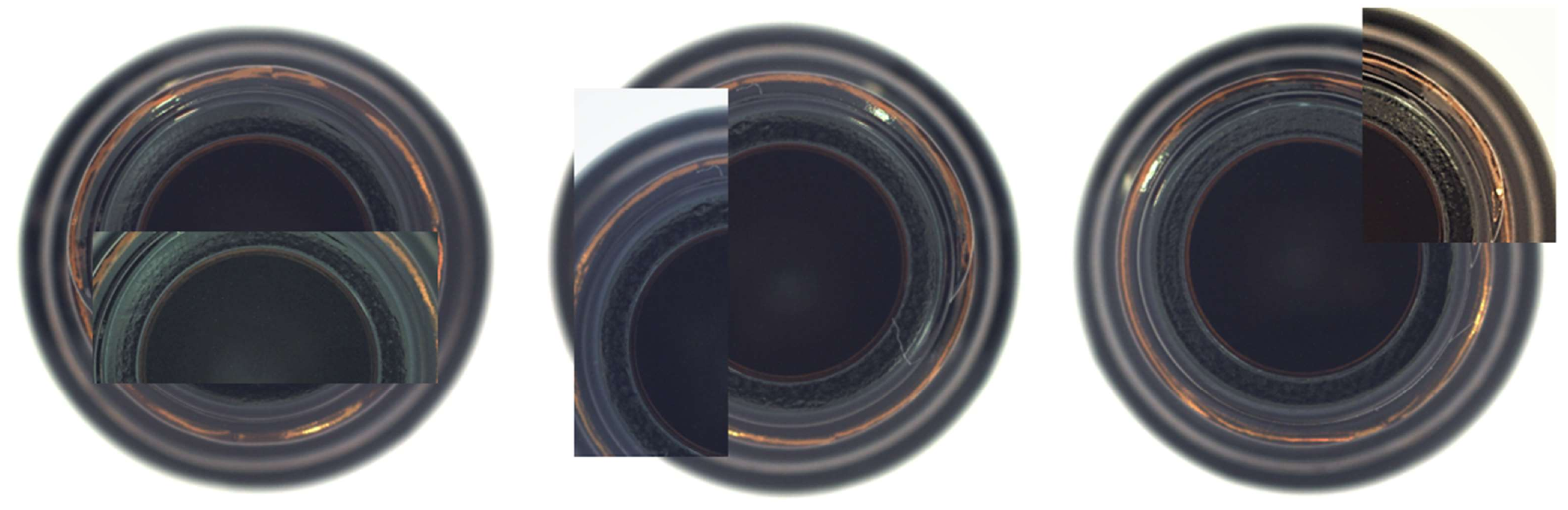}
    \caption{CutPaste-style processed images. An image patch is sampled from the full image, then color-jittered and pasted onto the original image.}
    \label{fig:CutPaste}
\end{figure}

We also examined the robustness of anomalous data not included in the test data.
From Fig.~\ref{fig:toy_problem}, we hypothesized that the generator learns the distribution of real-normal data, and the discriminator lays the discrimination boundary between high-density and low-density regions of normal data. The following two types of data were used for the evaluation: fake-anomalous data generated by $\sigma=5$ to $8$, and CutPaste-style processed images~\cite{li2021cutpaste} on normal data included in the test data.
An image patch was sampled from the full image with an area ratio of $20$ to $30\%$ and an aspect ratio between $3:1$ and $1:3$, then color-jittered and pasted onto the image.

As in the previous experiments, the discriminator was trained using fake-anomalous data generated by $\sigma=4$ in the Bottle category. All the anomalous data, generated by $\sigma=5$ to $8$ and CutPaste images, were classified as anomalous. Thus, these results suggest that the discriminator learns similar to one-class classification between normal data and the other data types.

\section{Discussion}
\label{sec:discussion}
The key finding of this study is that adding pseudo-anomalous data to training improves the anomaly detection performance of the discriminator, as shown in Table~\ref{table:ablation}.
Fake-anomalous data, one of the pseudo-anomalous data types, are generated from anomalous latent variables with high entropy. This method has fewer concerns about biases than adding out-of-distribution data to the training~\cite{kawachi2018complementary,hendrycks2018deep} or using prior knowledge to generate pseudo-anomalous data~\cite{li2021cutpaste}.

Furthermore, our proposed method uses only the discriminator for anomaly detection, whereas other image-based methods use multiple networks such as encoders, decoders, or both. Thus, the prediction time of our method is faster.

PatchCore~\cite{roth2021towards} is a state-of-the-art feature-based method that uses pre-trained models for images and cannot utilize other types of data. The same is true for other high-performance methods. By contrast, our proposed method can be directly applied to both images and features. Standard GANs have the potential to approximate any data distribution. ALGAN, an extension of standard GANs, has the same potential.

The evaluation in this study focused on image data. However, it would be interesting to see the performance of ALGAN for anomaly or novelty detection in other data types, such as signals~\cite{brophy2021generative} and text~\cite{DEROSA2021108098}. This is an important topic for future work.

\section{Conclusion}
\label{sec:conclusion}

We proposed a novel GAN-based anomaly detection method called ALGAN. The ALGAN generator provides pseudo-anomalous data as well as fake-normal data, by introducing anomalous states in the latent variable. The ALGAN discriminator distinguishes between the group of real-normal data and the group of fake-normal and pseudo-anomalous data.

The proposed method for generating pseudo-anomalous data can be applied to both images and feature vectors. We applied it to three anomaly detection benchmarks and demonstrated its high accuracy.

On MVTec-AD, ALGAN-image achieved more than 10\% higher average accuracy than conventional image-based methods, and ALGAN-feature exhibited comparable ability to the feature-based methods. On the COIL-100 dataset, ALGAN performed almost perfectly.

ALGAN exhibited remarkably fast predictions. Compared with methods trained on image data and features, ALGAN-image could predict 10.4 to 54.6 times faster while maintaining high performance, and ALGAN-feature could predict 1.3 to 2.2 times faster.

\appendix

\bibliography{aaai22}

\begin{thebibliography}{51}
\providecommand{\natexlab}[1]{#1}

\bibitem[{Ahmad et~al.(2017)Ahmad, Lavin, Purdy, and
  Agha}]{ahmad2017unsupervised}
Ahmad, S.; Lavin, A.; Purdy, S.; and Agha, Z. 2017.
\newblock Unsupervised real-time anomaly detection for streaming data.
\newblock \emph{Neurocomputing}, 262: 134--147.

\bibitem[{Ak{\c{c}}ay, Atapour-Abarghouei, and
  Breckon(2018)}]{akcay2018ganomaly}
Ak{\c{c}}ay, S.; Atapour-Abarghouei, A.; and Breckon, T.~P. 2018.
\newblock Ganomaly: Semi-supervised anomaly detection via adversarial training.
\newblock In \emph{Proc. Asian Conf. Comput. Vis. (ACCV)}, 622--637. Springer.

\bibitem[{Ak{\c{c}}ay, Atapour-Abarghouei, and Breckon(2019)}]{akccay2019skip}
Ak{\c{c}}ay, S.; Atapour-Abarghouei, A.; and Breckon, T.~P. 2019.
\newblock Skip-ganomaly: Skip connected and adversarially trained
  encoder-decoder anomaly detection.
\newblock In \emph{Proc. Int. Joint Conf. Neural Netw. (IJCNN)}, 1--8. IEEE.

\bibitem[{Andrews et~al.(2016)Andrews, Tanay, Morton, and
  Griffin}]{andrews2016transfer}
Andrews, J.; Tanay, T.; Morton, E.~J.; and Griffin, L.~D. 2016.
\newblock Transfer representation-learning for anomaly detection.
\newblock In \emph{ICML 2016, Anomaly Detection Workshop}.

\bibitem[{Bergman, Cohen, and Hoshen(2020)}]{bergman2020deep}
Bergman, L.; Cohen, N.; and Hoshen, Y. 2020.
\newblock Deep nearest neighbor anomaly detection.
\newblock \emph{arXiv preprint arXiv:2002.10445}.

\bibitem[{Bergmann et~al.(2019{\natexlab{a}})Bergmann, Fauser, Sattlegger, and
  Steger}]{bergmann2019mvtec}
Bergmann, P.; Fauser, M.; Sattlegger, D.; and Steger, C. 2019{\natexlab{a}}.
\newblock MVTec AD--A Comprehensive Real-World Dataset for Unsupervised Anomaly
  Detection.
\newblock In \emph{Proc. IEEE/CVF Conf. Comput. Vis. Pattern Recognit. (CVPR)},
  9592--9600. IEEE.

\bibitem[{Bergmann et~al.(2019{\natexlab{b}})Bergmann, L{\"{o}}we, Fauser,
  Sattlegger, and Steger}]{BergmannLFSS19}
Bergmann, P.; L{\"{o}}we, S.; Fauser, M.; Sattlegger, D.; and Steger, C.
  2019{\natexlab{b}}.
\newblock Improving Unsupervised Defect Segmentation by Applying Structural
  Similarity to Autoencoders.
\newblock In \emph{Proc. Int. Joint Conf. Comput. Vis., Imag. Comput. Graph.
  Theory App. (VISAPP)}, 372--380. INSTICC.
\newblock ISBN 978-989-758-354-4.

\bibitem[{Brophy et~al.(2021)Brophy, Wang, She, and
  Ward}]{brophy2021generative}
Brophy, E.; Wang, Z.; She, Q.; and Ward, T. 2021.
\newblock Generative adversarial networks in time series: A survey and
  taxonomy.
\newblock \emph{arXiv preprint arXiv:2107.11098}.

\bibitem[{Chalapathy and Chawla(2019)}]{chalapathy2019deep}
Chalapathy, R.; and Chawla, S. 2019.
\newblock Deep learning for anomaly detection: A survey.
\newblock \emph{arXiv preprint arXiv:1901.03407}.

\bibitem[{Chalapathy, Menon, and
  Chawla(2018)}]{chalapathy2018oneclassneuralnet}
Chalapathy, R.; Menon, A.~K.; and Chawla, S. 2018.
\newblock Anomaly detection using one-class neural networks.
\newblock \emph{arXiv preprint arXiv:1802.06360}.

\bibitem[{Chandola, Banerjee, and Kumar(2009)}]{Chandola_etal_2009}
Chandola, V.; Banerjee, A.; and Kumar, V. 2009.
\newblock Anomaly Detection: A Survey.
\newblock \emph{ACM Computing Surveys}, 41(3).

\bibitem[{Chatillon and Ballester(2020)}]{chatillon2020history}
Chatillon, P.; and Ballester, C. 2020.
\newblock History-based anomaly detector: an adversarial approach to anomaly
  detection.
\newblock In \emph{Proc. SAI Intelligent Syst. Conf.}, 761--776. Springer.

\bibitem[{Cohen and Hoshen(2020)}]{cohen2020sub}
Cohen, N.; and Hoshen, Y. 2020.
\newblock Sub-image anomaly detection with deep pyramid correspondences.
\newblock \emph{arXiv preprint arXiv:2005.02357}.

\bibitem[{{de Rosa} and Papa(2021)}]{DEROSA2021108098}
{de Rosa}, G.~H.; and Papa, J.~P. 2021.
\newblock A survey on text generation using generative adversarial networks.
\newblock \emph{Pattern Recognition}, 119: 108098.

\bibitem[{Defard et~al.(2021)Defard, Setkov, Loesch, and
  Audigier}]{defard2021padim}
Defard, T.; Setkov, A.; Loesch, A.; and Audigier, R. 2021.
\newblock PaDiM: A patch distribution modeling framework for anomaly detection
  and localization.
\newblock In \emph{Pattern Recognit. ICPR Int. Workshops Challenges}, 475--489.
  Springer.

\bibitem[{Deng et~al.(2009)Deng, Dong, Socher, Li, Li, and
  Fei-Fei}]{deng2009imagenet}
Deng, J.; Dong, W.; Socher, R.; Li, L.-J.; Li, K.; and Fei-Fei, L. 2009.
\newblock Imagenet: A large-scale hierarchical image database.
\newblock In \emph{Proc. IEEE Conf. Comput. Vis. Pattern Recognit. (CVPR)},
  248--255. IEEE.

\bibitem[{Dinh, Sohl{-}Dickstein, and Bengio(2017)}]{DBLP:conf/iclr/DinhSB17}
Dinh, L.; Sohl{-}Dickstein, J.; and Bengio, S. 2017.
\newblock Density estimation using Real {NVP}.
\newblock In \emph{Int. Conf. Learning Represent. (ICLR)}.

\bibitem[{Goodfellow et~al.(2014)Goodfellow, Pouget-Abadie, Mirza, Xu,
  Warde-Farley, Ozair, Courville, and Bengio}]{goodfellow2014generative}
Goodfellow, I.; Pouget-Abadie, J.; Mirza, M.; Xu, B.; Warde-Farley, D.; Ozair,
  S.; Courville, A.; and Bengio, Y. 2014.
\newblock Generative Adversarial Nets.
\newblock In \emph{Proc. Adv. Neural Inf. Process. Syst.}, 2672--2680.

\bibitem[{He et~al.(2016)He, Zhang, Ren, and Sun}]{he2016deep}
He, K.; Zhang, X.; Ren, S.; and Sun, J. 2016.
\newblock Deep residual learning for image recognition.
\newblock In \emph{Proc. IEEE Conf. Comput. Vis. Pattern Recognit. (CVPR)},
  770--778. IEEE.

\bibitem[{Hendrycks, Mazeika, and Dietterich(2018)}]{hendrycks2018deep}
Hendrycks, D.; Mazeika, M.; and Dietterich, T. 2018.
\newblock Deep Anomaly Detection with Outlier Exposure.
\newblock In \emph{Int. Conf. Learning Represent. (ICLR)}.

\bibitem[{Huang, Qiu, and Yuan(2020)}]{huang2020surface}
Huang, Y.; Qiu, C.; and Yuan, K. 2020.
\newblock Surface defect saliency of magnetic tile.
\newblock \emph{The Visual Computer}, 36(1): 85--96.

\bibitem[{Ioffe and Szegedy(2015)}]{ioffe2015batch}
Ioffe, S.; and Szegedy, C. 2015.
\newblock Batch Normalization: Accelerating Deep Network Training by Reducing
  Internal Covariate Shift.
\newblock In \emph{Proc. Int. Conf. Mach. Learn. (ICML)}, 448--456. PMLR.

\bibitem[{Johnson, Douze, and J{\'e}gou(2021)}]{johnson2019billionfaiss}
Johnson, J.; Douze, M.; and J{\'e}gou, H. 2021.
\newblock Billion-scale similarity search with gpus.
\newblock \emph{IEEE Transactions on Big Data}, 7(3): 535--547.

\bibitem[{Kawachi, Koizumi, and Harada(2018)}]{kawachi2018complementary}
Kawachi, Y.; Koizumi, Y.; and Harada, N. 2018.
\newblock Complementary set variational autoencoder for supervised anomaly
  detection.
\newblock In \emph{Proc. IEEE Int. Conf. Acoust. Speech Signal Process.
  (ICASSP)}, 2366--2370. IEEE.

\bibitem[{Kingma and Ba(2015)}]{kingma:adam}
Kingma, D.~P.; and Ba, J. 2015.
\newblock Adam: A method for stochastic optimization.
\newblock In \emph{Int. Conf. Learning Represent. (ICLR)}.

\bibitem[{Krizhevsky, Sutskever, and Hinton(2012)}]{NIPS2012_c399862d}
Krizhevsky, A.; Sutskever, I.; and Hinton, G.~E. 2012.
\newblock ImageNet Classification with Deep Convolutional Neural Networks.
\newblock In \emph{Proc. Adv. Neural Inf. Process. Syst.}, volume~25,
  1097--1105.

\bibitem[{Li et~al.(2021)Li, Sohn, Yoon, and Pfister}]{li2021cutpaste}
Li, C.-L.; Sohn, K.; Yoon, J.; and Pfister, T. 2021.
\newblock CutPaste: Self-Supervised Learning for Anomaly Detection and
  Localization.
\newblock In \emph{Proc. IEEE/CVF Conf. Comput. Vis. Pattern Recognit. (CVPR)},
  9664--9674. IEEE.

\bibitem[{Liu et~al.(2020)Liu, Ouyang, Wang, Fieguth, Chen, Liu, and
  Pietik{\"a}inen}]{liu2020deepobjectdetection}
Liu, L.; Ouyang, W.; Wang, X.; Fieguth, P.; Chen, J.; Liu, X.; and
  Pietik{\"a}inen, M. 2020.
\newblock Deep learning for generic object detection: A survey.
\newblock \emph{International Journal of Computer Vision}, 128(2): 261--318.

\bibitem[{Menghani(2021)}]{menghani2021efficient}
Menghani, G. 2021.
\newblock Efficient Deep Learning: A Survey on Making Deep Learning Models
  Smaller, Faster, and Better.
\newblock \emph{arXiv preprint arXiv:2106.08962}.

\bibitem[{Miyato et~al.(2018)Miyato, Kataoka, Koyama, and
  Yoshida}]{miyato2018spectral}
Miyato, T.; Kataoka, T.; Koyama, M.; and Yoshida, Y. 2018.
\newblock Spectral Normalization for Generative Adversarial Networks.
\newblock In \emph{Int. Conf. Learning Represent. (ICLR)}.

\bibitem[{Nene et~al.(1996)Nene, Nayar, Murase et~al.}]{nene1996columbia}
Nene, S.~A.; Nayar, S.~K.; Murase, H.; et~al. 1996.
\newblock Columbia Object Image Library (COIL-100).

\bibitem[{Pang et~al.(2021)Pang, Shen, Cao, and
  Hengel}]{pang2021deepanomreview}
Pang, G.; Shen, C.; Cao, L.; and Hengel, A. V.~D. 2021.
\newblock Deep Learning for Anomaly Detection: A Review.
\newblock \emph{ACM Computing Surveys}, 54(2).

\bibitem[{Pedregosa et~al.(2011)Pedregosa, Varoquaux, Gramfort, Michel,
  Thirion, Grisel, Blondel, Prettenhofer, Weiss, Dubourg, Vanderplas, Passos,
  Cournapeau, Brucher, Perrot, and Duchesnay}]{scikit-learn}
Pedregosa, F.; Varoquaux, G.; Gramfort, A.; Michel, V.; Thirion, B.; Grisel,
  O.; Blondel, M.; Prettenhofer, P.; Weiss, R.; Dubourg, V.; Vanderplas, J.;
  Passos, A.; Cournapeau, D.; Brucher, M.; Perrot, M.; and Duchesnay, E. 2011.
\newblock Scikit-learn: Machine Learning in {P}ython.
\newblock \emph{Journal of Machine Learning Research}, 12: 2825--2830.

\bibitem[{Perales~G\'{o}mez et~al.(2019)Perales~G\'{o}mez,
  Fern\'{a}ndez~Maim\'{o}, Huertas~Celdr\'{a}n, Garc\'{i}a~Clemente,
  Cadenas~Sarmiento, Del Canto~Masa, and M\'{e}ndez~Nistal}]{8926471}
Perales~G\'{o}mez, A.~L.; Fern\'{a}ndez~Maim\'{o}, L.; Huertas~Celdr\'{a}n, A.;
  Garc\'{i}a~Clemente, F.~J.; Cadenas~Sarmiento, C.; Del Canto~Masa, C.~J.; and
  M\'{e}ndez~Nistal, R. 2019.
\newblock On the Generation of Anomaly Detection Datasets in Industrial Control
  Systems.
\newblock \emph{IEEE Access}, 7: 177460--177473.

\bibitem[{Pourreza et~al.(2021)Pourreza, Mohammadi, Khaki, Bouindour, Snoussi,
  and Sabokrou}]{pourreza2021g2d}
Pourreza, M.; Mohammadi, B.; Khaki, M.; Bouindour, S.; Snoussi, H.; and
  Sabokrou, M. 2021.
\newblock G2d: Generate to detect anomaly.
\newblock In \emph{Proc. IEEE/CVF Wint. Conf. App. Comput. Vis. (WACV)},
  2003--2012. IEEE.

\bibitem[{Radford, Metz, and Chintala(2016)}]{DBLP:journals/corr/RadfordMC15}
Radford, A.; Metz, L.; and Chintala, S. 2016.
\newblock Unsupervised Representation Learning with Deep Convolutional
  Generative Adversarial Networks.
\newblock In \emph{Int. Conf. Learning Represent. (ICLR)}.

\bibitem[{Rippel, Mertens, and Merhof(2021)}]{rippel2021modeling}
Rippel, O.; Mertens, P.; and Merhof, D. 2021.
\newblock Modeling the distribution of normal data in pre-trained deep features
  for anomaly detection.
\newblock In \emph{Proc. Int. Conf. Pattern Recognit. (ICPR)}, 6726--6733.
  IEEE.

\bibitem[{Rodda and Erothi(2016)}]{rodda2016classimbalance}
Rodda, S.; and Erothi, U. S.~R. 2016.
\newblock Class imbalance problem in the network intrusion detection systems.
\newblock In \emph{Proc. Int. Conf. Electrical, Electronics, Optim. Techniq.
  (ICEEOT)}, 2685--2688. IEEE.

\bibitem[{Roth et~al.(2021)Roth, Pemula, Zepeda, Sch{\"o}lkopf, Brox, and
  Gehler}]{roth2021towards}
Roth, K.; Pemula, L.; Zepeda, J.; Sch{\"o}lkopf, B.; Brox, T.; and Gehler, P.
  2021.
\newblock Towards Total Recall in Industrial Anomaly Detection.
\newblock \emph{arXiv preprint arXiv:2106.08265}.

\bibitem[{Rudolph, Wandt, and Rosenhahn(2021)}]{RudWan2021}
Rudolph, M.; Wandt, B.; and Rosenhahn, B. 2021.
\newblock Same same but differnet: Semi-supervised defect detection with
  normalizing flows.
\newblock In \emph{Proc. IEEE/CVF Wint. Conf. App. Comput. Vis. (WACV)},
  1907--1916. IEEE.

\bibitem[{Ruff et~al.(2021)Ruff, Kauffmann, Vandermeulen, Montavon, Samek,
  Kloft, Dietterich, and Müller}]{ruff2020unifying}
Ruff, L.; Kauffmann, J.~R.; Vandermeulen, R.~A.; Montavon, G.; Samek, W.;
  Kloft, M.; Dietterich, T.~G.; and Müller, K.-R. 2021.
\newblock A Unifying Review of Deep and Shallow Anomaly Detection.
\newblock \emph{Proceedings of the IEEE}, 109(5): 756--795.

\bibitem[{Sabokrou et~al.(2015)Sabokrou, Fathy, Hoseini, and
  Klette}]{sabokrou2015real}
Sabokrou, M.; Fathy, M.; Hoseini, M.; and Klette, R. 2015.
\newblock Real-time anomaly detection and localization in crowded scenes.
\newblock In \emph{Proc. IEEE Conf. Comput. Vis. Pattern Recognit. Workshops
  (CVPRW)}, 56--62. IEEE.

\bibitem[{Sabokrou et~al.(2018)Sabokrou, Khalooei, Fathy, and
  Adeli}]{sabokrou2018adversarially}
Sabokrou, M.; Khalooei, M.; Fathy, M.; and Adeli, E. 2018.
\newblock Adversarially Learned One-Class Classifier for Novelty Detection.
\newblock In \emph{Proc. IEEE/CVF Conf. Comput. Vis. Pattern Recognit. (CVPR)},
  3379--3388. IEEE.

\bibitem[{Schlegl et~al.(2017)Schlegl, Seeb{\"o}ck, Waldstein, Schmidt-Erfurth,
  and Langs}]{schlegl2017unsupervised}
Schlegl, T.; Seeb{\"o}ck, P.; Waldstein, S.~M.; Schmidt-Erfurth, U.; and Langs,
  G. 2017.
\newblock Unsupervised anomaly detection with generative adversarial networks
  to guide marker discovery.
\newblock In \emph{Conf. Inf. Process. Med. Imaging (IPMI)}, 146--157.
  Springer.

\bibitem[{Sener and Savarese(2018)}]{sener2018activekcentergreedy}
Sener, O.; and Savarese, S. 2018.
\newblock Active Learning for Convolutional Neural Networks: A Core-Set
  Approach.
\newblock In \emph{Int. Conf. Learning Represent. (ICLR)}.

\bibitem[{Simonyan and Zisserman(2015)}]{Simonyan15}
Simonyan, K.; and Zisserman, A. 2015.
\newblock Very Deep Convolutional Networks for Large-Scale Image Recognition.
\newblock In \emph{Int. Conf. Learning Represent. (ICLR)}.

\bibitem[{Sinha et~al.(2020)Sinha, Zhang, Goyal, Bengio, Larochelle, and
  Odena}]{sinha2020smallrandomprojection}
Sinha, S.; Zhang, H.; Goyal, A.; Bengio, Y.; Larochelle, H.; and Odena, A.
  2020.
\newblock Small-gan: Speeding up gan training using core-sets.
\newblock In \emph{Proc. Int. Conf. Mach. Learn. (ICML)}, 9005--9015. PMLR.

\bibitem[{Tan and Le(2019)}]{tan2019efficientnet}
Tan, M.; and Le, Q. 2019.
\newblock Efficientnet: Rethinking model scaling for convolutional neural
  networks.
\newblock In \emph{Int. Conf. Mach. Learn. (ICML)}, 6105--6114. PMLR.

\bibitem[{Zagoruyko and Komodakis(2016)}]{zagoruyko2016wide}
Zagoruyko, S.; and Komodakis, N. 2016.
\newblock Wide Residual Networks.
\newblock In \emph{British Mach. Vis. Conf. (BMVC)}. BMVC.

\bibitem[{Zaheer et~al.(2020)Zaheer, Lee, Astrid, and Lee}]{zaheer2020old}
Zaheer, M.~Z.; Lee, J.-h.; Astrid, M.; and Lee, S.-I. 2020.
\newblock Old is Gold: Redefining the Adversarially Learned One-Class
  Classifier Training Paradigm.
\newblock In \emph{Proc. IEEE/CVF Conf. Comput. Vis. Pattern Recognit. (CVPR)},
  14183--14193. IEEE.

\bibitem[{Zenati et~al.(2018)Zenati, Foo, Lecouat, Manek, and
  Chandrasekhar}]{zenati2018efficient}
Zenati, H.; Foo, C.~S.; Lecouat, B.; Manek, G.; and Chandrasekhar, V.~R. 2018.
\newblock Efficient gan-based anomaly detection.
\newblock \emph{arXiv preprint arXiv:1802.06222}.

\end{thebibliography}

\end{document}